# Iterative Framework For Data Augmentation Of Segmented Fingerprints

João Leonardo Harres Dall Agnol*, Wesley Augusto de Bona*, Érick Oliveira Rodrigues*,
Luiz Fernando Puttow Southier*, Jefferson Oliva*, Marcelo Filipak†,
Dalcimar Casanova*
*Federal University of Technology (UTFPR)
Pato Branco, Parana Brazil
Email: {joaoleonardoagnol@alunos.utfpr.edu.br, wesleybona@alunos.utfpr.edu.br, erickrodrigues@utfpr.edu.br,
luizsouthier@utfpr.edu.br, jeffersonoliva@utfpr.edu.br, dalcimar@utfpr.edu.br}
†InfantID
Curitiba, Parana, Brazil
Email: filipak@natosafe.com

***Abstract*—Infant biometrics presents unique challenges due to the physiological differences between infants and adults, compounded by the scarcity of available data for research that limits the development of robust matching systems. This paper proposes a novel data augmentation method that uses iterative techniques to generate diverse variants of segmented fingerprints by inducing errors in a convolutional neural network trained to extract fingerprint ridges and valleys. Experiments on real infant fingerprints demonstrate the method's effectiveness in expanding fingerprint variability, with augmentations exhibiting significant fluctuations in minutiae counts while still retaining visual similarity to the originals. The study also highlights the method's customizable nature for applying varying levels of changes to fingerprint segmentations. Future research includes training segmentation and matching neural networks using datasets augmented by the proposed framework.**

***Index Terms*—Augmentation, Data, Fingerprint, Infant, Segmentation.**

## I. Introduction

Infant biometrics remains a relatively unexplored field, having received limited attention from both commercial solutions and academic research, despite the interest of international agencies [1], [2]. Biometric systems often encounter challenges with infants due to significant differences in anatomy compared to adults. In fingerprint biometrics specifically, infants' ridge/valley structure is notably thinner and more closely spaced due to the reduced dimensions of their fingers [2], [3], making it challenging to distinguish unique formations such as minutiae and other patterns. The deformation of children's fingerprints during capture is also a known obstacle, prompting the development of non-contact scanner alternatives [3], [4]. Ridge and valley segmentation is particularly affected by these characteristics because, unlike region of interest segmentation, the structural details of the fingerprint are extracted to a binarized image, meaning that noise and deformation of the fingerprint may significantly impact the output of these methods. Additionally, drastic aging effects have been observed, potentially limiting the time frame in which different captures of infants can be matched [5].

Given these factors, capturing infant prints is inherently more challenging than capturing adults, as obtaining high-quality images — with enough detail to enable segmentation of their thin ridges and valleys — may require the use of costly high-resolution (and possibly non-contact) scanners [3]. In this context, Deep Learning research emerges as a promising avenue for overcoming finger biometric challenges, as intelligent methods can be trained to identify infant fingerprint ridges and valley patterns, whereas traditional methods falter due to image quality or other complications. Nevertheless, acquiring sufficient data becomes a crucial aspect of any Deep Learning-oriented research, as these networks frequently require large amounts of examples during training [6]. Thus, one of the main hindrances of the infant recognition problem is the lack of availability of large and representative infant datasets [1].

Data augmentation methods are often used to address data scarcity. For fingerprints, these techniques typically involve conventional transformations like patch croppings, rotations, or introducing artificial occlusions to create several variants from a single original image [7]–[9]. Recently, Neural Networks have been used to generate synthetic samples of adult fingerprints [10], but this also requires large amounts of real fingerprints for effective training. This paper proposes a novel data augmentation method that emphasizes segmented (pre-processed) images rather than captured prints, thereby eliminating the need to replicate texture and contrast. By disregarding these patterns and focusing on minutiae, the method enables the generation of fingerprint variants from a single original image, enhancing the variability of the dataset without producing entirely artificial fingerprints. This more subtle approach to data augmentation aims to mimic the variability observed across multiple samples of the same individual.

## II. Related Works

While Deep Learning holds promise across various domains, its efficacy is fundamentally reliant on the availability and quality of training data. As noted by [6] and [9], the application of Deep Learning techniques to medical images is often problematic due to data scarcity and how often ground-truth annotations are non-representative. This presents a notable issue because Deep Learning Neural Networks strive to exhibit robustness, meaning they should be capable of generalizing to unseen examples, but achieving this capability often demands large amounts of representative data [6].

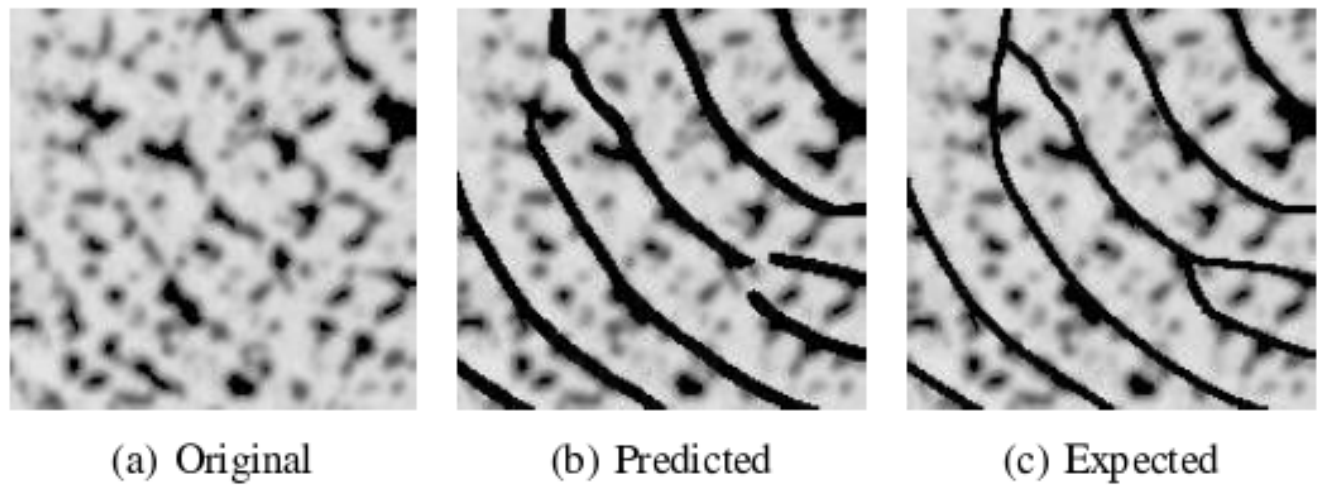

Fig. 1: Segmentation errors. For easier visualization, both predicted (automated) and expected (manually drawn by a specialist) segmentation were overlaid to the original image

In fingerprint biometrics, data augmentation serves to artificially expand datasets or simulate image capture anomalies. To enhance data diversity, conventional geometric transformations like cropping, rotations, and flips have been proven to be effective [7], [8]. Additionally, [11] outline a variety of operations, primarily localized cropping, aimed at replicating diverse forms of occlusion errors encountered during capture. More recently, Neural Networks have been employed to generate entirely new fingerprints [10]. However, as is the case for traditional Deep Learning, training these networks for robust performance also necessitates access to large and diverse datasets [12], which are often lacking in infant biometrics. Circumventing the issue of data availability, this study introduces a customizable data augmentation method that explores faults in intelligent ridge/valley segmentation methods to generate numerous variations from a single fingerprint segmentation.

## III. Proposed Augmentation Method

As previously covered in Section I, it is possible to use segmentation networks to extract the valleys and ridges of fingerprints. However, despite their capabilities these networks still struggle to achieve satisfactory segmentation, particularly when applied in children's biometrics. Errors are prevalent in such scenarios, which the comparison depicted in Figure 1 illustrates.

Segmentation mistakes often occur due to the lack of definition in the captured image (Figure 1a), making it difficult to distinguish between valleys and ridges. For example, in Figure 1, the automated segmentation (Figure 1b) fails to recognize two bifurcation minutiae present in the manually segmented ground truth (Figure 1c). This study utilizes the segmentation errors of a Convolutional Neural Network (CNN) from a third-party partner [13] to create segmentation variants. The CNN, originally trained to segment ridges and valleys of high-resolution infant fingerprints, produces binarized black-and-white images containing only the fingerprint valleys. The valleys are segmented in white, and background in black. However, for easier visualization, the illustrations included in this study were inverted (black valleys, white background).

By exclusively working with segmented fingerprints, the augmentation process avoids the complexities of replicating contrast and texture features in captured prints. This streamlined approach simplifies augmentation but limits it to creating variants within the segmented domain. However, in certain scenarios (such as segmented fingerprints matching) this trade off may prove advantageous, as it allows the generation of diverse variants, enriching the dataset and facilitating robust training of biometric recognition systems. The proposed data augmentation method, illustrated by Figure 2, comprises an iterative framework where previous segmentation attempts of a fingerprint are successively overlaid onto the captured image. This process relies on two user-defined arguments: (i) alpha ($\alpha$), which regulates the opacity of the overlay, and (ii) the number of repetitions ($n$), which determines the number of overlays performed.

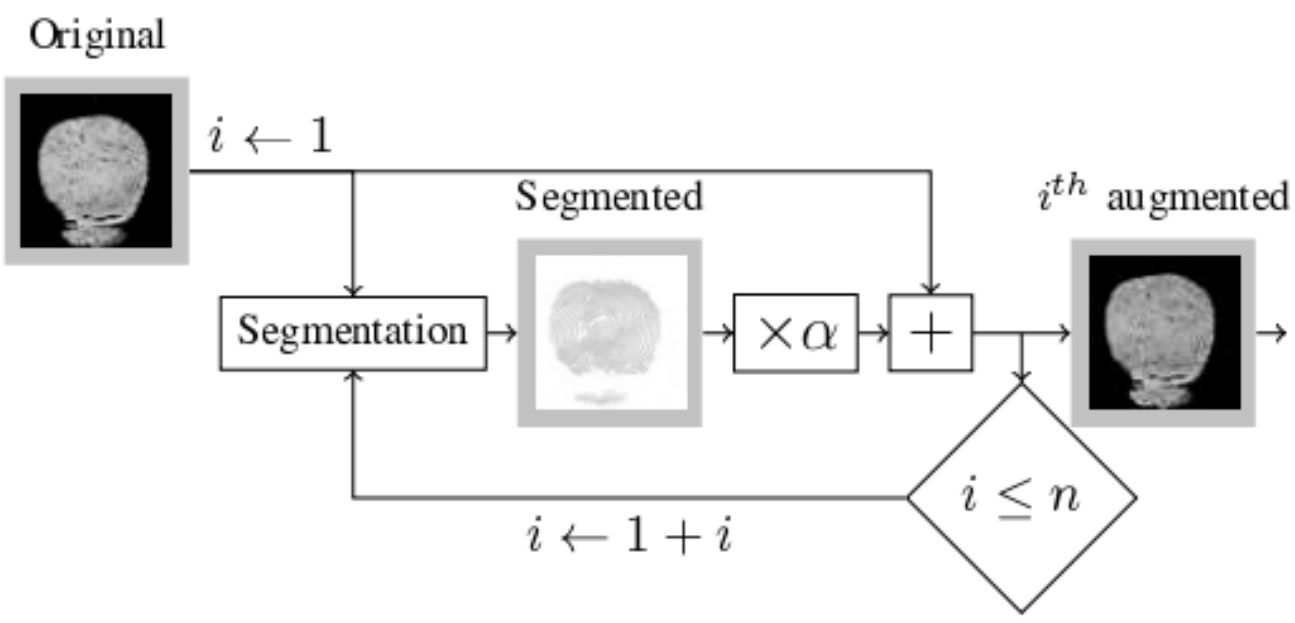

Fig. 2: Iterative augmentation pipeline

Processing begins as a fingerprint image is received as input:

- At $i = 1$, the CNN predicts the valley segmentation for the input image;
- The CNN output is then overlaid onto the input image using an $\alpha$ parameter (ranging from 0% for fully transparent to 100% for fully opaque) to control the intensity of the overlay;
- This combined image, containing the original fingerprint and the segmented valleys, serves as input for the next iteration, $i + 1$;

In subsequent iterations, the CNN receives this overlay, where the valley segmentation partially obscures the original image. The degree of information loss is controlled by $\alpha$: an overlay with $\alpha$ set to 100% fully replaces the original image with the valley segmentation output, while an $\alpha$ of 50% produces a dimmed fingerprint with the valleys clearly segmented on top. By continuously re-utilizing these modified outputs in successive iterations, the predicted valleys are transformed, with each subsequent prediction being slightly modified as the model extends previously identified lines.

As shown in Figure 3, this process connects and shifts valley segments, altering the minutiae map while preserving the fingerprint's overall structure.

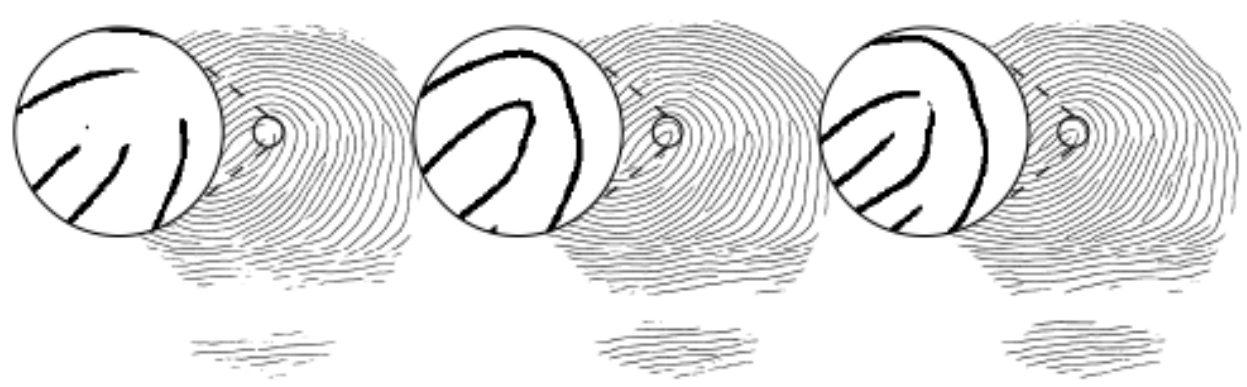

Fig. 3: Comparison between different augmentation iterations. From left to right: 1st iteration, 15th iteration, 30th iteration

In summary, this study presents a customizable data augmentation framework that utilizes image overlays to exploit segmentation errors, enabling the generation of multiple segmentation variants that can enrich fingerprint datasets independent of the subjects' age. The extent to which this method enables data augmentation and the impact of the input (user-defined) parameters is further investigated in Section IV.

## IV. Experimental results and discussion

Two experiments are conducted to evaluate the effectiveness of the proposed method in enhancing fingerprint segmentation. First, each fingerprint's minutiae map is analyzed, with changes in minutiae count compared as both the $\alpha$ parameter and the iteration threshold vary. Subsequently, an image similarity comparison is executed to investigate the behavior of the $\alpha$ parameter, assessing how different tuning levels impact the augmented outputs.

These experiments utilize a dataset comprising five high-resolution (3200 ppi) infant fingerprints collected by [13], featuring samples from infants aged between 0 and 36 hours. All samples are shown in Figure 4.

### *A. Minutiae map comparison*

Fingerprint matchers typically rely on minutiae-based approaches, making these features essential for accurate authentication [14]. Therefore, deformations to the minutiae map that alter the location and classification of these critical points can effectively create variants of the same fingerprint.

Table I compares the number of minutiae found during various iterations of the proposed method with high (50%) and low (5%) $\alpha$ values. These minutiae were extracted using an image processing method [15] that identifies valley endings and bifurcations, excluding border minutiae, in a skeletonized segmentation (one-pixel-wide structures).

Based on the findings in Table I, high $\alpha$ values enable the proposed method to significantly alter the minutiae count of infant fingerprints, regardless of the original segmentation. For instance, Finger B exhibits the highest number of minutiae at 97, with reductions ranging from 10% (1st iteration, 88 minutiae) to almost 54% (15th iteration, 57 minutiae). In contrast, Finger E initially has 54 minutiae but reveals 98 in the first augmentation, marking an increase of 81%. These fluctuations—whether reductions or increases in minutiae—demonstrate the effectiveness of the minutiae map deformations.

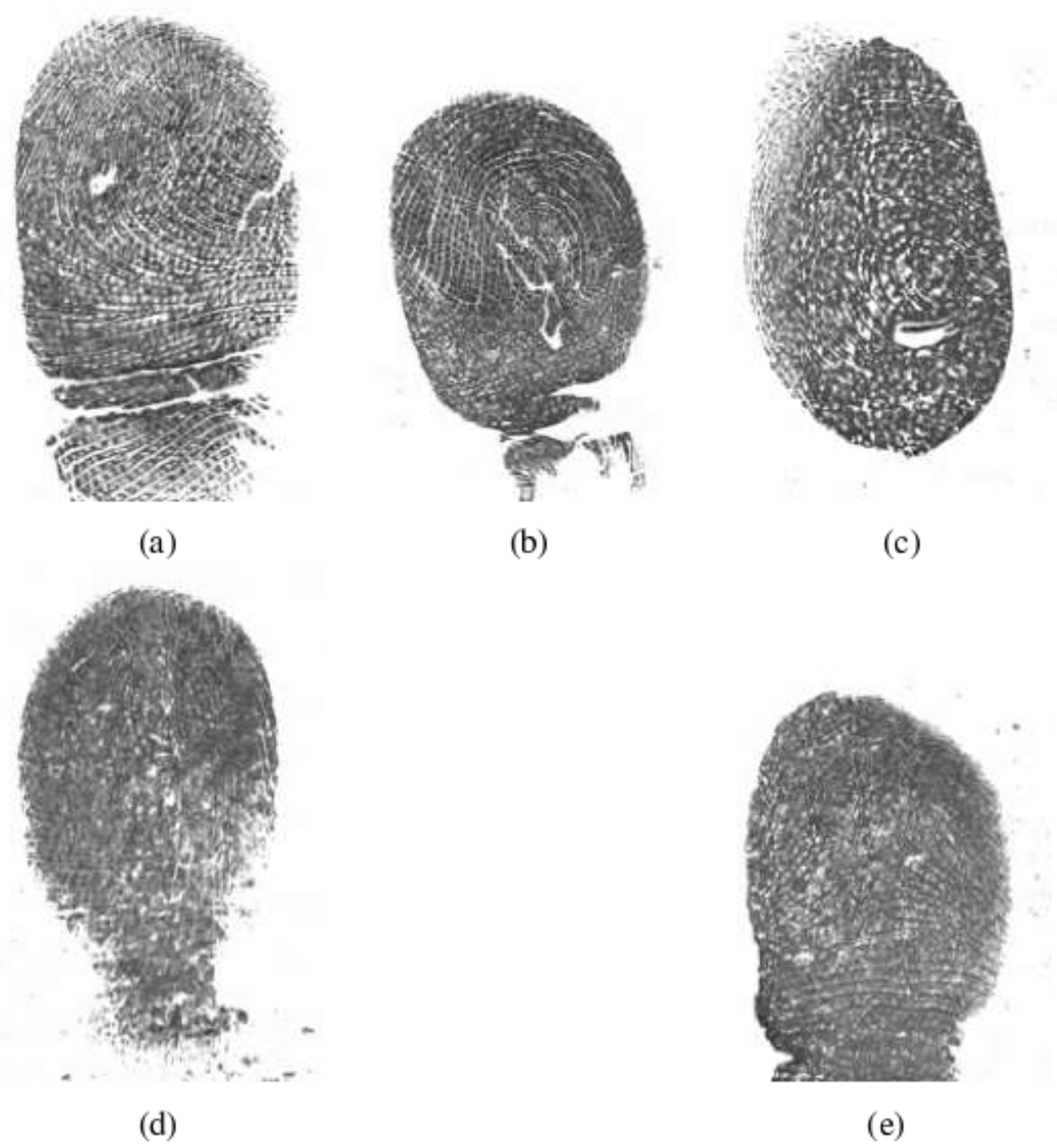

(a) (b) (c) (d) (e)

Fig. 4: Infant fingerprint samples

Tab. I: Minutiae count comparison between original image, low and high $\alpha$ augmentations

| Finger | A | B | C | D | E |
|---|---|---|---|---|---|
| **Reference** | 57 | 97 | 69 | 76 | 54 |
| **1st iter. (high $\alpha$)** | 63 | 88 | 53 | 67 | 98 |
| **15th iter. (high $\alpha$)** | 89 | 45 | 28 | 40 | 48 |
| **30th iter. (high $\alpha$)** | 64 | 57 | 32 | 40 | 38 |
| **1st iter. (low $\alpha$)** | 60 | 95 | 69 | 74 | 99 |
| **15th iter. (low $\alpha$)** | 48 | 98 | 61 | 42 | 68 |
| **30th iter. (low $\alpha$)** | 49 | 96 | 62 | 37 | 72 |

[a]Minutiae counts from iterative augmentations.

It is important to note that the proposed augmentation method is capable of creating new structures connections that can erase line endings and form new bifurcations. Although the minutiae count may remain unchanged, the minutiae map is still altered due to changes in the classification of the minutiae. Figure 5 illustrates this dynamic: Finger D's reference initially shows many incomplete lines identified as fingerprint borders. These lines were treated as terminations during minutiae extraction in the first iteration but were extended by the 15th iteration, ultimately resulting in a new bifurcation by the 30th iteration. While three minutiae are still present in the highlighted area, the minutiae map now consists of two line endings followed by a bifurcation, rather than a cluster of line endings.

The occurrence of new line connections that eliminate minutiae is seemingly common, as shown in Table I, where high $\alpha$ iterations for Finger C reduce the minutiae count from 69 (reference) to only 28 (15th iteration). However, with further increases in iterations, new minutiae are generated,

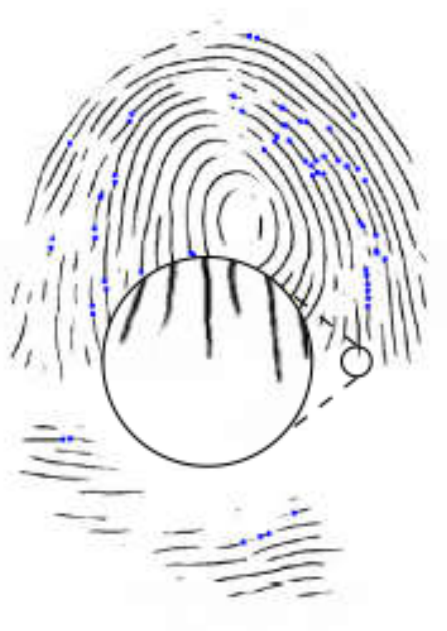
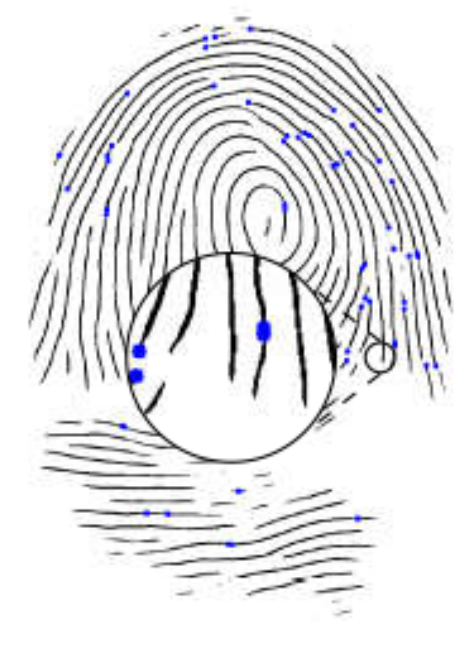
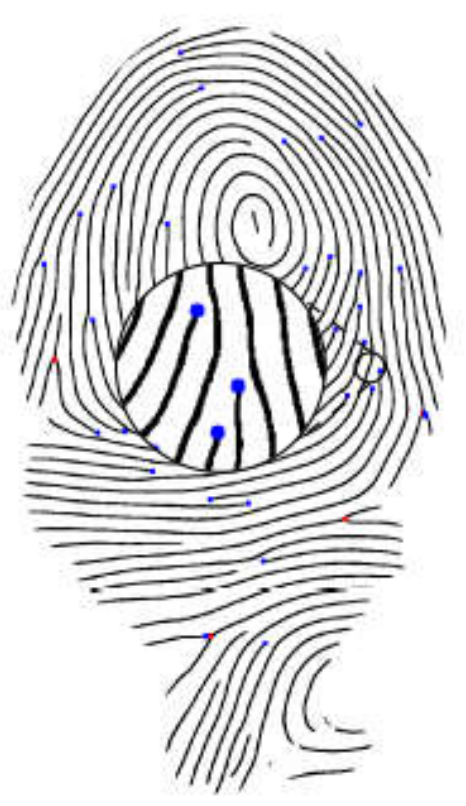
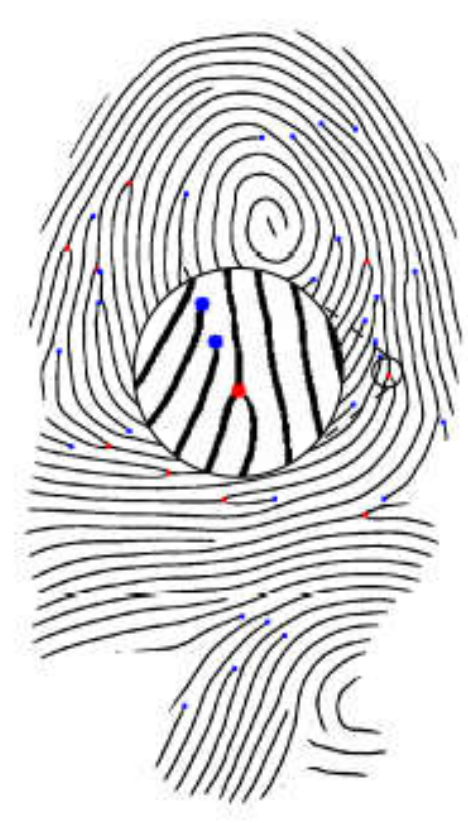

Fig. 5: Finger D's localized minutiae comparison for high $\alpha$ augmentation, with endings colored in blue and bifurcations in red. From left to right: reference, 1st iteration, 15th iteration, 30th iteration

raising the count to 32 by the 30th iteration.

Regarding finer augmentation using low $\alpha$ values, Table I demonstrates slight variations in minutiae count. In Finger B, for example, the number of minutiae only fluctuates by 2 (from 97 to 95 or 98). However, instances like Finger E show that abrupt changes still occur even with softer parameters, as the minutiae count almost doubles between the reference (54 minutiae) and a single augmentation iteration (99 minutiae). Nonetheless, the comparison between Finger B's reference and the 30th iteration, illustrated in Figure 6, depicts gradual changes. A localized comparison over 30 iterations shows no alteration in the ending region, but the ending itself slightly shifts to the left, creating a new bifurcation.

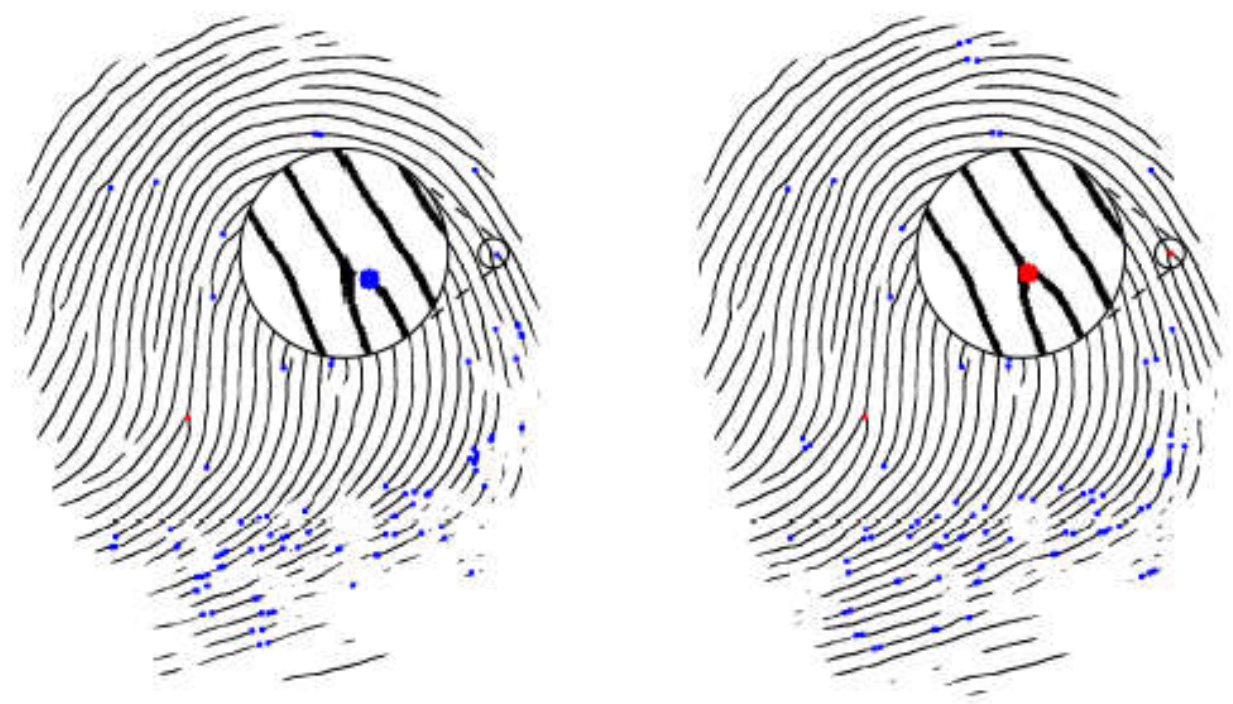

Fig. 6: Comparison between Finger B's reference (left) and the 30th iteration (right) using a low $\alpha$. Endings are colored in blue, and bifurcations in red

Regardless of parameter configurations, these findings strongly suggest that the proposed method can generate a wide range of minutiae variations based on a single image, producing fingerprint variants to augment data. This approach diversifies fingerprint datasets through minutiae alterations rather than simply increasing the number of samples with traditional image-processing operations, such as rotation and cropping. This distinct characteristic could be advantageous in developing robust infant matching systems, as it provides similar-looking fingerprints with subtly or markedly different minutiae maps, mimicking the deformations to which children's fingerprints are particularly susceptible.

### B. *Parameter tuning*

As evidenced by Subsection IV-A, the proposed method can generate both profound and subtle variants of segmented images. By adjusting both parameters, $\alpha$ and $n$, users can control the extent of iterations and the abruptness of changes between them. Thus, this section explores how the similarity between original images and iterations fluctuates over the course of 30 iterations as $\alpha$ changes. To gauge the similarity of samples, the Dice coefficient was utilized. Originally developed by [16] and [17] to compare species overlap, the Dice coefficient's equation (Equation 1) computes the similarity between sets $A$ and $B$ using set operations, making it similarly efficient for computing image similarity, including finger biometrics [18], [19]. In this section, comparison is performed at the segmentation level, considering only differences between valleys.

$$Dice(A, B) = \frac{2 \times |A \cap B|}{|A| + |B|} \tag{1}$$

Figure 7 showcases multiple data augmentations of Finger B with different $\alpha$ values, illustrating how the Dice score changes over the course of 30 iterations. In Figure 7a, the Dice scores for low $\alpha$ values ranging from 1% to 10%, increasing by 1% per set, are depicted. Upon examining Figure 7a, it becomes apparent that with slight adjustments to the $\alpha$ parameter, the proposed method exhibits a generally linear behavior. Specifically, the similarity between the reference and each iteration decreases proportionally to the $\alpha$ applied. This linear trend suggests a predictable relationship between the degree of augmentation and the resulting similarity scores.

However, these minor adjustments do not accumulate significantly, as all curves exhibit an elbow, indicating that changes stabilize after a few iterations. Figure 7b supports this, showing

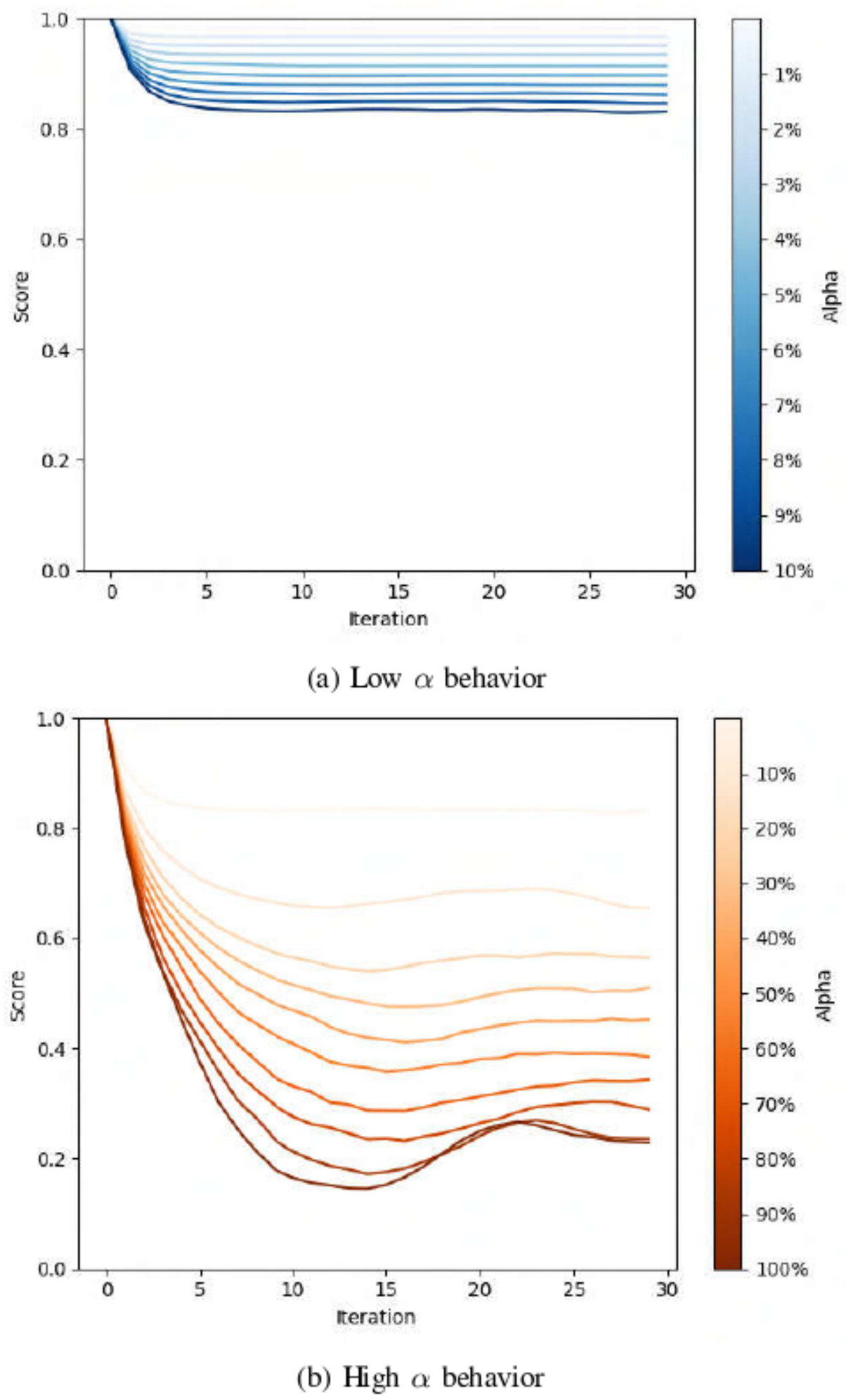

Fig. 7: Proposed method behavior with various $\alpha$ values. Low values range from 1% to 10%, and high values range from 10% to 100%

Dice scores for augmentations with high $\alpha$ values (10% to 100%) and abrupt increases (10% per set). Both Figure 7a and Figure 7b demonstrate that even at the maximum $\alpha$ value, the method cannot generate a completely new fingerprint. This means that even distant iterations of a high $\alpha$ setup still resemble the original segmentation, indicating that the proposed method is only capable of creating variants (and not completely artificial fingerprints), ensuring some matching potential across all iterations and possibly allowing the simulation of multiple captures of the same individual.

Another key observation highlighted by Figure 7b is that this elbow can be shifted, as high $\alpha$ values tend to stabilize in later iterations. This shift can be severe in the case of values higher than 70%, whose curves exhibit fluctuations that indicate continuous changes in the similarity of iterations.

## V. Conclusion

This paper addresses the challenges in biometric research due to the scarcity of infant datasets, which undermines the recognition of children's fingerprints. To overcome this limitation and reduce data acquisition costs, a novel data augmentation method is proposed. Utilizing iterative techniques, this method generates detailed minutiae variations of segmented fingerprints, capable of not only augmenting the number of samples through traditional transformations but also enhancing the diversity of individuals while requiring only a single sample as input. Experiments on a real infant fingerprint dataset demonstrate that the proposed method significantly expands the variability of segmented fingerprints through substantial minutiae map deformations, capable of both adding and removing large quantities of minutiae. An initial quantitative analysis using the Dice coefficient suggests that the method can significantly alter the fingerprint's structure, regardless of changes to minutiae types and locations. However, further studies may require additional evaluation metrics to fully assess how well the augmentation preserves the original fingerprint's similarity.

Future research directions includes training segmentation and matching networks using augmented datasets generated by the proposed method. Additionally, efforts could be directed towards the development of a novel image quality index focused on fingerprint segmentation differences to better explore how automated segmentation adds or removes ridges and valleys in relation to their ground-truth.

## VI. Acknowledgements

We sincerely thank InfantID, CAPES (Finance Code 001), CNPq, Araucária Foundation, and FINEP, for their financial support.